\documentclass[10pt,twocolumn,letterpaper]{article}

\usepackage{cvpr}
\usepackage{times}
\usepackage{epsfig}
\usepackage{graphicx}
\usepackage{amsmath}
\usepackage{amssymb}
\usepackage{multirow}
\cvprfinalcopy % *** Uncomment this line for the final submission

\def\cvprPaperID{753} % *** Enter the CVPR Paper ID here

\ifcvprfinal\fi
\begin{document}

%%%%%%%%% TITLE
\title{Deep MANTA: A Coarse-to-fine Many-Task Network for joint 2D and 3D vehicle analysis from monocular image}

\author{Florian Chabot$^{1}$ , Mohamed Chaouch$^{1}$ , Jaonary Rabarisoa$^{1}$ , C\'{e}line Teuli\`{e}re$^{2}$ , Thierry Chateau$^{2}$ \\
$^{1}$ CEA-LIST Vision and Content Engineering Laboratory, $^{2}$ Pascal Institute, Blaise Pascal University\\
%%Institution1 address\\
{\tt\small $^{1}$\{florian.chabot, mohamed.chaouch, jaonary.rabarisoa\}@cea.fr} \\
{\tt\small $^{2}$\{celine.teuliere, thierry.chateau\}@univ-bpclermont.fr‎ }
%\and
%Celine Teuliere$^{\dagger}$, Thierry Chateau$^{\dagger}$\\
}

\maketitle
%\thispagestyle{empty}

%%%%%%%%% ABSTRACT
\maketitle
\begin{abstract}
In this paper, we present a novel approach, called Deep MANTA (Deep Many-Tasks), for many-task vehicle analysis from a given image. A robust convolutional network is introduced for simultaneous vehicle detection, part localization, visibility characterization and 3D dimension estimation. Its architecture is based on a new coarse-to-fine object proposal that boosts the vehicle detection. Moreover, the Deep MANTA network is able to localize vehicle parts even if these parts are not visible. In the inference, the network's outputs are used by a real time robust pose estimation algorithm for fine orientation estimation and 3D vehicle localization. We show in experiments that our method outperforms monocular state-of-the-art approaches on vehicle detection, orientation and 3D location tasks on the very challenging KITTI benchmark.

\end{abstract}

%------------------------------------------------------------------------- 
\section{Introduction}
\label{sec:intro}
\thispagestyle{empty}
%Over the last years, traffic scene analysis has been improved thanks to deep learning approaches and thus paves the way to multiple applications, especially, autonomous driving. Impressive recent work in 2D object detection  \cite{faster,rcnn,fast} already provides important information related to scenes content but do not allow to describe objects in the 3D real world scene.
%In this paper, we are interested in both 2D and 3D vehicle analysis from monocular images in the context of self-driving car. It is an active research field because most current cars are equipped with a single camera. Given a single image, the proposed approach provides accurate vehicle detection, vehicle part localization, vehicle part visibility, fine orientation, 3D localization and 3D template. Figure \ref{fig:outputs} illustrates the outputs of our approach. In practical application, 3D vehicle localization and orientation jointly used with temporal description are essential to recover vehicles speeds and directions and thus predict critical situations. Moreover, knowing vehicle part location allows to describe the vehicle in a finer way. To give an example, correctly localizing high lights is important to interpret vehicle direction indicators. Finally, the characterization of the visibility of vehicle parts is also very useful in real cases. It allows to know if the vehicle is hidden by other vehicles or environment obstacles.

\begin{figure}[t]
\includegraphics[width=8.6cm]{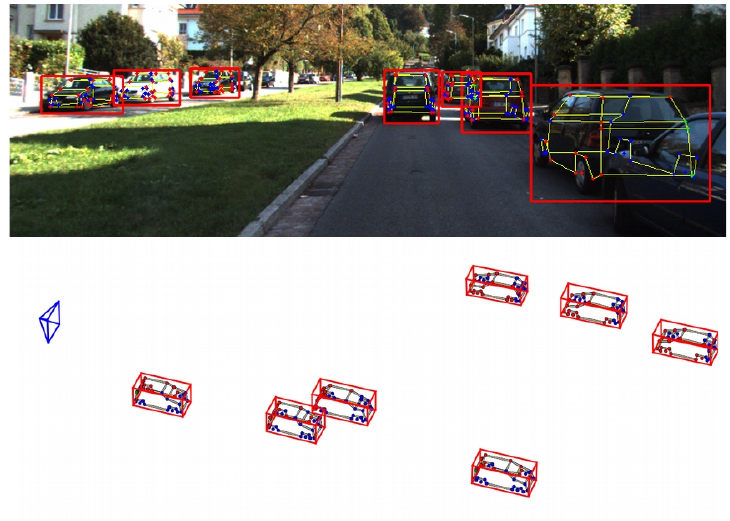}
\vspace*{-5mm}
\centering
\caption{System outputs. \textit{Top}: 2D vehicle bounding boxes, vehicle part localization and part visibility. In this example, red dots correspond to visible parts, green dots to occluded parts and blue dots to self-occluded parts. \textit{Bottom}: 3D vehicle bounding box localization and 3D vehicle part localization. The camera is represented in blue.}
\vspace{-3mm}

\label{fig:outputs}
\end{figure}

Over the last years, traffic scene analysis has been improved thanks to deep learning approaches which paves the way to multiple applications, especially, autonomous driving. Impressive recent work in 2D object detection~\cite{faster,rcnn,fast} already provides important information related to scenes content but does not yet allow to describe objects in the 3D real world scene. In this paper, we are interested in both 2D and 3D vehicle analysis from monocular images in the context of self-driving cars. This is a relevant research field because currently most cars are equipped with a single camera. For an autonomously driving vehicle, it is essential to understand the traffic and predict critical situations based on the information extracted from the image of the scene. For the recovery of speed and direction of the surrounding cars, 3D vehicle localization and orientation jointly used with temporal description are necessary. Additionally, for proper traffic understanding it is important to describe surrounding vehicles in a fine way. For example, correct localization of high lights is required to interpret vehicle direction indicators, for which knowledge of the exact location of vehicle parts is needed. Finally, for interpretation of the overall scene the characterization of the visibility of vehicle parts needs also to be obtained. Thus it will be known  if a vehicle is hidden by other vehicles or environment obstacles. Here we propose an approach that, given a single image, provides accurate vehicle detections, vehicle part localization, vehicle part visibility, fine orientation, 3D localization and 3D template (3D dimension). Figure~\ref{fig:outputs} illustrates the outputs of our approach. 

Our first contribution is to encode 3D vehicle information using characteristic points of vehicles. The underlying idea is that 3D vehicle information can be recovered using monocular images because vehicles are rigid objects with well known geometry. Our approach localizes vehicle parts even if these parts are hidden due to occlusion, truncation or self-occlusion in the image. These parts are found using regression instead of using a part detector. In this way, the approach predicts the position of hidden parts which are essential for robust 3D information recovering. We use a 3D vehicle dataset composed of 3D meshes with real dimensions. Several vertices are annotated for each 3D model. These 3D points correspond to vehicle parts (such as wheels, headlights, etc) and define a 3D shape for each 3D model. The main idea of the approach is to recover the projection of these 3D points (2D shape) in the input image for each detected vehicle. Then, the best corresponding 3D model for each detection box is chosen. 2D/3D matching is performed between 2D shapes and selected 3D shapes to recover vehicle orientation and 3D location.

The second contribution is the introduction of the Deep Coarse-to-fine Many-Task Convolutional Neural Network called Deep MANTA. This network outputs accurate 2D vehicle bounding boxes, 2D shapes, part visibility and 3D vehicle templates. Its architecture contains several originalities. Firstly, inspired by the Region proposal network~\cite{faster}, the MANTA model is able to propose coarse 2D bounding boxes which are then iteratively refined, by multi-pass forward, to provide accurate scored 2D detections. Secondly, this network is based on the many-task concept. That means that the same feature vector can be used to predict many tasks. We optimize in the same time six tasks: region proposal, detection, 2D box regression, part localization, part visibility and 3D template prediction.

The last contribution is related to the training dataset. Deep neural networks require many samples and labels to be efficiently learned. Furthermore, it is very fastidious and almost impossible to annotate manually vehicle parts which are not visible. For this purpose, we propose a semi-automatic annotation process using 3D models to generate labels on real images for the Deep MANTA training. Labels from 3D models (geometry information, visibility, etc) are automatically projected onto real images providing a large training dataset without labour-intensive annotation work. 

In the next section, related work is reviewed. The section~\ref{sec:method} explains the proposed model. Finally, we show that our approach outperforms monocular state-of-the-art methods related to vehicle detection, orientation and 3D localization on the very challenging KITTI dataset~\cite{kitti}.

\begin{figure*}[ht]
\includegraphics[width=17.2cm]{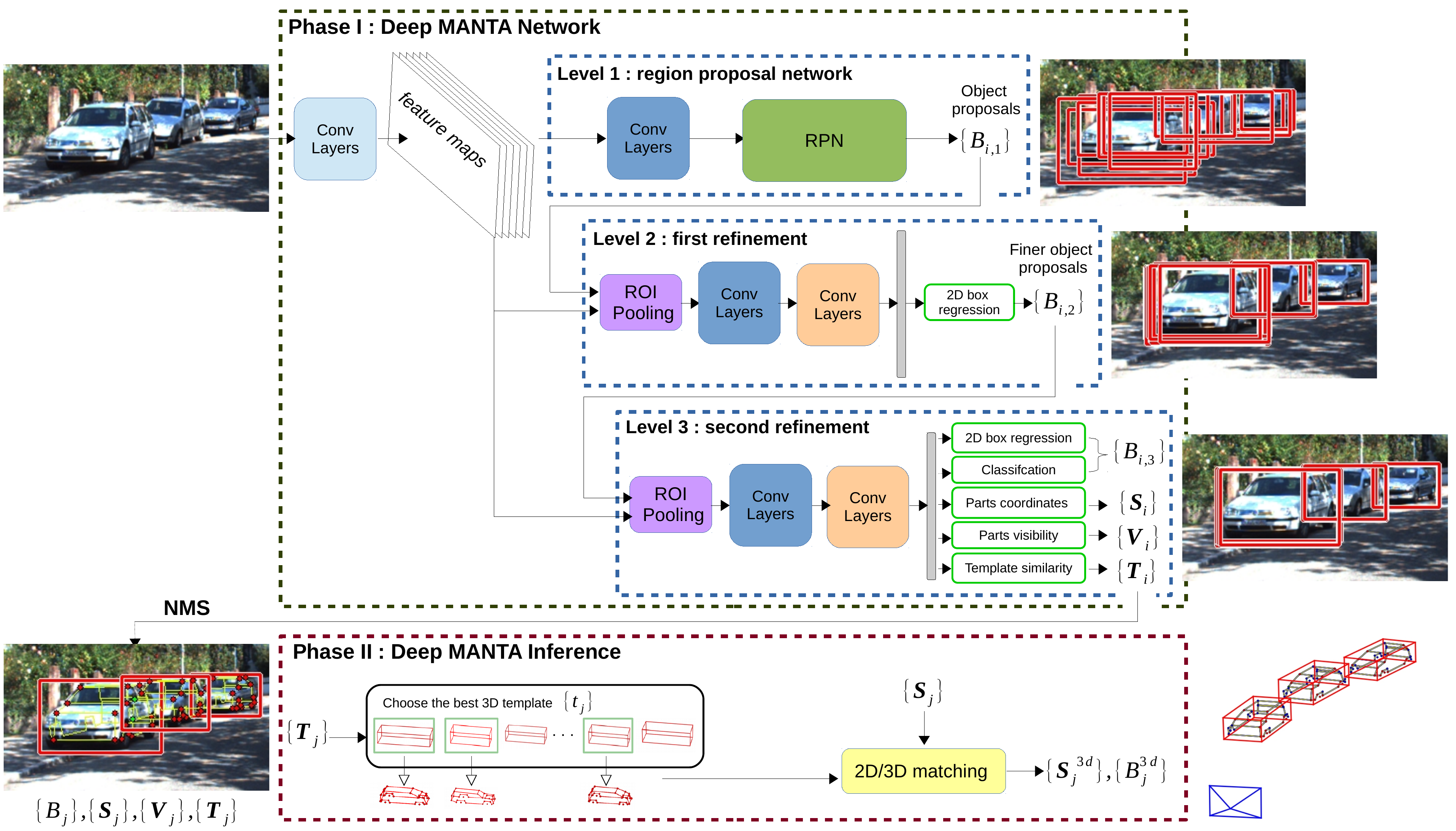}
\centering
\vspace{1mm}
\caption{Overview of the Deep MANTA approach. The entire input image is forwarded inside the Deep MANTA network. Conv layers with the same color share the same weights. Moreover, these three convolutional blocks correspond to the split of existing CNN architecture. The network provides object proposals $\{B_{i,1}\}$ which are iteratively refined ($\{B_{i,2}\}$ and then the final detection set $\{B_{i,3}\}$). 2D part coordinates $\{\bold{S}_i\}$, part visibility $\{\bold{V}_i\}$ and template similarity $\{\bold{T}_i\}$ are associated to the final set of detected vehicle $\{B_{i,3}\}$. A non-maximum suppression (NMS) is then performed. It removes redundant detections and provides the new set $\{B_{j},\bold{S}_j,\bold{V}_j,\bold{T}_j\}$. Using these outputs, the inference step allows to choose the best corresponding 3D template using template similarity $\bold{T}_j$ and then performs 2D/3D pose computation using the associated 3D shape.}
\label{fig:shapercnn}
%\vspace{-3mm}

\end{figure*}

\section{Related work}
\label{sec:rw}

Object analysis is a well studied topic and we divide it into two main categories: 2D object detection/coarse pose estimation and 3D object detection/fine pose estimation.

\textbf{2D Object detection and coarse pose estimation.} There are two ways to perform 2D object detection. The first one is the standard sliding window scheme used in many detection systems as~\cite{Felzenszwalb10,overfeat}. The second one is the 2D object proposal based methods~\cite{rcnn,fast,Uijlings13,Carreira10constrainedparametric,APBMM2014}. The goal of object proposal methods is to propose several boxes with high objectness confidence score. These proposals are then given to a detector which is able to classify objects and background. The main advantage of object proposal methods is the processing time because that considerably reduces the search space. In parallel, Deep Convolutional Neural Networks (CNN) have proven their effectiveness in many computer vision fields such as object classification~\cite{gn, resnet,bird,Ranzato}, object detection~\cite{rcnn,fast,faster} and scene segmentation~\cite{seg1,seg2}. Thus, the success of object proposal methods as well as CNN, leads people to directly learn \textit{Region Proposal Networks} (RPN) sharing weights with the down-stream detection network~\cite{faster,sdp, subcnn, hypernet}. RPN provides strong objectness confidence regions of interest computed on deep feature maps. Experiments show that this kind of method increases detection accuracy. The proposed approach uses the RPN framework but uses several steps of 2D bounding box refinement to significantly increase object detection performance.
2D object detection is often associated with pose estimation and many methods address the two issues. They generally divide the viewing sphere in several bins to learn multi-class models where each bin corresponds to a class~\cite{Savarese, Savarese2, Lepetit, Schmid, Pepik}. These approaches allow to get coarse information on objects and do not provide continuous viewpoint estimation. 

\textbf{3D Object detection and fine pose estimation}. To go further than 2D reasoning, several approaches are designed to detect vehicles in 3D space and are able to give a detailed 3D object representation. A part of them consists in fitting 3D models~\cite{Ikea,Pepik2, Aubry, Lim}, active shape model~\cite{Zia1,Zia2,Zia3,eccv14_fit,Savarese3} or predicting 3D voxel patterns~\cite{3dvp} to recover the exact 3D pose and detailed object representation. These methods generally use an initialization step providing the 2D bounding box and the coarse viewpoint information. More recently, people have proposed to use 3D object proposals generated while using monocular images~\cite{mono3d} or disparity maps~\cite{3dop}. In these approaches, 3D object proposals are projected in 2D bounding boxes and given to a CNN based detector which jointly predicts the class of the object proposal and the object fine orientation (using angle regression). In the proposed approach, vehicle fine orientation estimation is found using a robust 2D/3D vehicle part matching: the 2D/3D pose matrix is computed using all vehicle parts (visible or hidden) in contrast to other methods such as~\cite{Zia1, Zia2, Zia3, eccv14_fit} which focus on visible parts. That clearly increases the precision of orientation estimation.

\section{Deep MANTA approach}  
\label{sec:method}

In this section, we describe the proposed approach for 2D/3D vehicle analysis from monocular images. Our system has two main steps. First, the input image is passed through the Deep MANTA network that outputs 2D scored bounding boxes, associated vehicle geometry (vehicle part coordinates, 3D template similarity) and part visibility properties. The Deep MANTA network architecture is detailed in the section \ref{Deep MANTA network}. The second step is the inference which uses Deep MANTA outputs and a 3D vehicle dataset to recover 3D orientations and locations. This step is detailed in the section  \ref{Inference}. 
In this method, we use a dataset of 3D shapes and one of 3D templates. These two datasets encode the variability of vehicles in terms of dimension, type, and shape. These datasets are presented in the section \ref{3dshapetemplate}. In the section \ref{Vehicle shape model}, we define the adopted 2D/3D vehicle model for a given vehicle in a monocular image.

\subsection{3D shape and template datasets}
\label{3dshapetemplate}

\begin{figure}[ht]
\includegraphics[width=8.6cm]{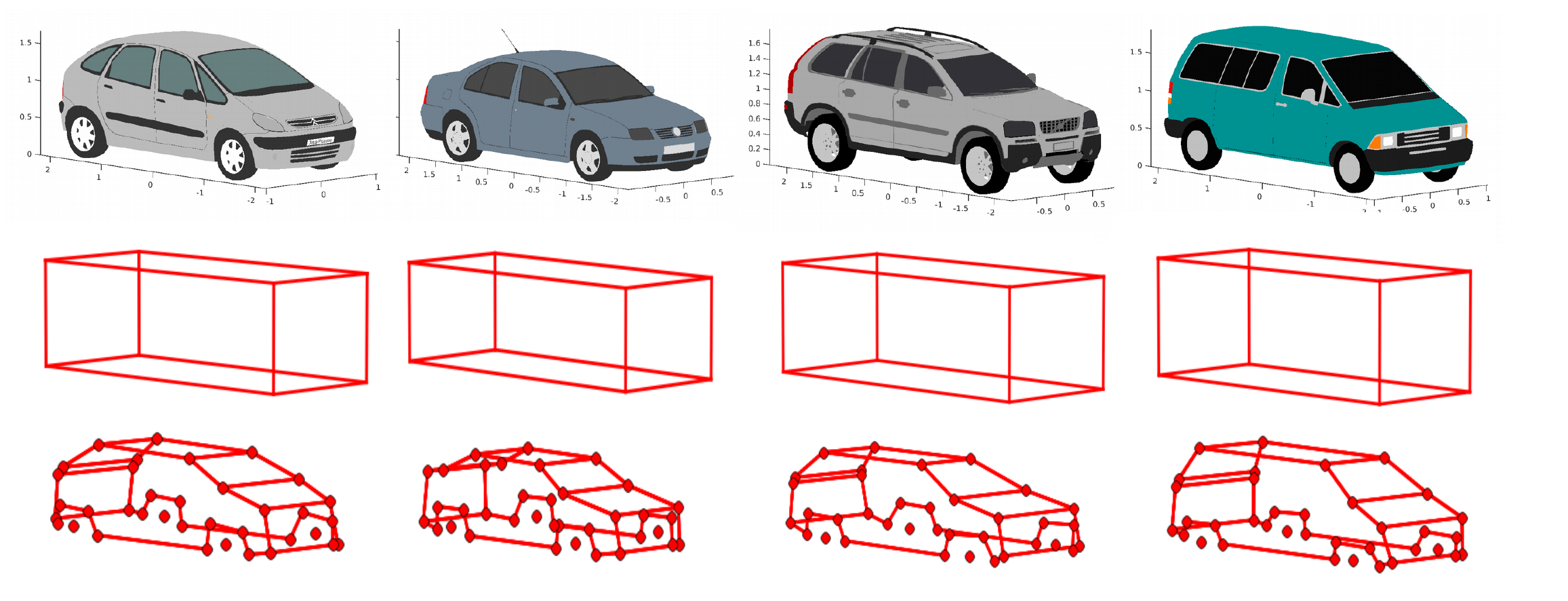}
\vspace{-3mm}
\caption{Some examples from the 3D template and 3D shape dataset. Each 3D model $m$ (first line) is associated to a 3D template $\bar{t}_m^{3d}$ (second line) and a 3D shape  $\bar{\bold{S}}_m^{3d}$ (third line). The 3D shape corresponds to manually annotated vertices. }
\label{fig:3ddataset}
\end{figure}

We use a dataset of $M$ 3D models corresponding to several types of vehicles (Sedan, SUV, etc). For each 3D model $m$, we annotate $N$ vertices (called 3D parts). These parts correspond to relevant vehicle regions. For one 3D model $m$, we denote its 3D shape aligned in canonical view as $\bar{\bold{S}}_m^{3d}=(p_1,p_2,..,p_N)$ with $p_k = (x_k,y_k,z_k)$ corresponding to the 3D coordinate of the $k^{th}$ part. The 3D template (\textit{i.e} 3D dimension) associated to the 3D model $m$ is defined as~ $\bar{t}^{3D}_m~=~(w_m,h_m,l_m)$ where $w_m$, $h_m$, $l_m$ are the width, the height and the length of the 3D model respectively. Figure~\ref{fig:3ddataset} shows some examples from the 3D shape dataset $\{\bar{\bold{S}}_m^{3d}\}_{m \in \{1,.., M\}}$ and the 3D template dataset $\{\bar{t}^{3d}_m\}_{m \in \{1,.., M\}}$. 

\subsection{2D/3D vehicle model}
\label{Vehicle shape model}

\begin{figure}[ht]
\includegraphics[width=7.0cm]{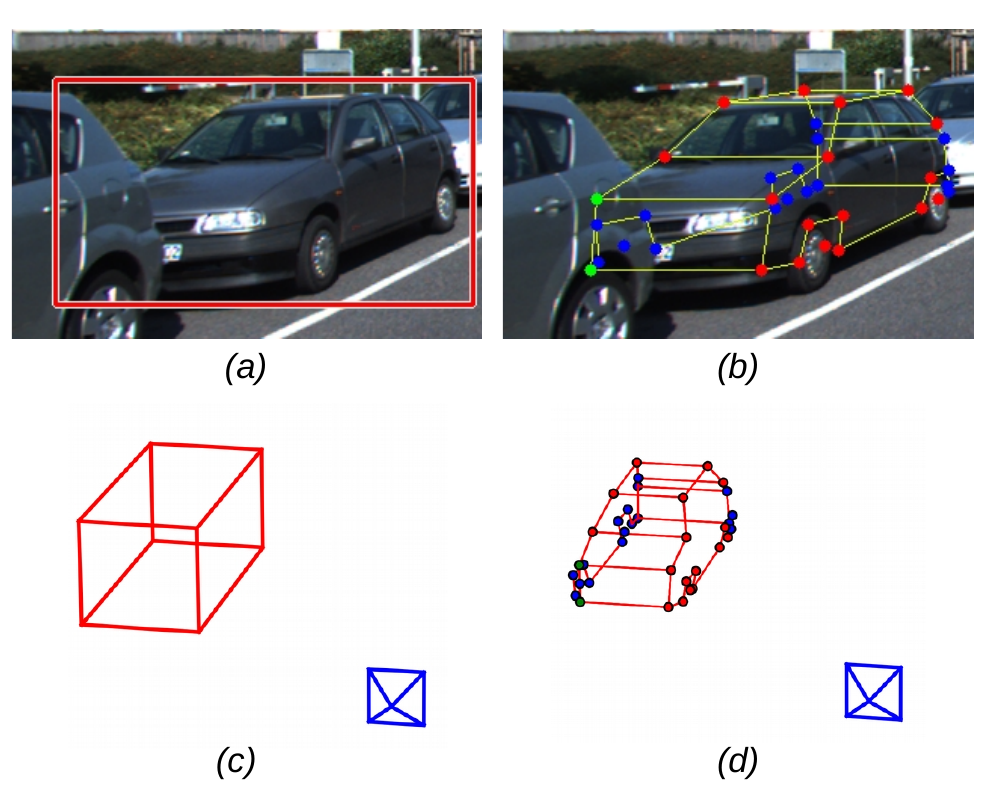}
\centering
\vspace{-1mm}
\caption{Example of one 2D/3D vehicle model. \textit{(a)} the bounding box $B$, \textit{(b)} 2D part coordinates $\bold{S}$ and part visibility $\bold{V}$: visible parts (red), occluded parts (green) and self-occluded parts (blue). \textit{(c)} the 3D bounding box $B^{3d}$ and \textit{(d)} the associated 3D shape $\bold{S}^{3d}$.}
\label{fig:vis}
\end{figure}

We represent each vehicle in a monocular image with a 2D/3D model. It is formally defined by the following attributes:
\begin{equation}
(B, B^{3d}, \bold{S}, \bold{S}^{3d}, \bold{V}) \notag
\end{equation}
$B=(c_x,c_y,w,h)$ is the 2D vehicle bounding box in the image where $(c_x,c_y)$ is the center and $(w,h)$ represents the width and the height respectively.~$B^{3d}~=~(c_x,c_y,c_z,\theta, t)$ is the 3D bounding box characterized by its 3D center $(c_x,c_y,c_z)$, its orientation $\theta$ and its 3D template~$t~=~(w,h,l)$ corresponding to its 3D real size. $\bold{S}=\{q_k = (u_k,v_k) \}_{k \in \{1,..,N\}}$ is the vehicle 2D part coordinates in the image. $\bold{S}^{3d}=\{p_k =(x_k,y_k,z_k) \}_{k \in \{1,..,N\}}$ is the vehicle 3D part coordinates in the 3D real word coordinate system. $\bold{V} = \{v_k\}_{k \in \{1,..,N\}}$ is the part visibility vector where $v_k$ denotes the visibility class of the $k^{th}$ part. Four classes of visibility are defined: (1) visible if the part is observed in the image, (2) occluded if the part is occluded by another object, (3) self-occluded if the part is occluded by the vehicle and (4) truncated if the part is out of the image. Figure~\ref{fig:vis} shows an example of a 2D/3D vehicle model.  

\subsection{Deep MANTA Network}
\label{Deep MANTA network}

The Deep MANTA network is designed to detect vehicles using a coarse-to-fine bounding box proposal as well as to output other finer attributes such as vehicle part localization, part visibility, and template similarity.

\textbf{Coarse-to-fine forward}. Given an entire input image, the network returns a first set of $K$ object proposals $\bold{B}_1 = \{B_{i,1}\}_{i \in \{1, .., K\}}$ as the region proposal network proposed by~\cite{faster}. These regions are then extracted from a feature map and pooled to a fixed size using ROI Pooling introduced by~\cite{fast}. Extracted regions are forwarded in a network (sharing some weights with the first level) and refined by offset transformations. A second set of $K$ objects $\bold{B}_2 = \{B_{i,2}\}_{i \in \{1, .., K\}}$ is proposed. This operation is repeated one last time to provide the final set of bounding box $\bold{B}_3$. These three levels of refinement are illustrated in Figure~\ref{fig:shapercnn}. This procedure differs than Faster-RCNN~\cite{faster} in that our iterative refinement steps overcome the constraints of large object scale variations and provide more accurate detection. Furthermore, in our approach, ROI pooled regions are extracted on the first convolution feature maps for keeping high resolution to detect hard vehicles.

\textbf{Many-task prediction}. The Deep MANTA architecture outputs a final bounding box set $\bold{B}_3 = \{B_{i,3}\}_{i \in \{1, .., K\}}$. For each bounding box $B_{i,3}$, the MANTA network also returns all 2D vehicle part coordinates $\bold{S}_i$, part visibility $\bold{V}_i$ and 3D template similarity $\bold{T}_i$. The template similarity vector $\bold{T}_i$ is defined as $\bold{T}_i = \{r_m\}_{m \in \{1,.., M\}}$.~$r_m~=~(r_x,r_y,r_z)$ corresponds to the three scaling factors to apply on the 3D template $\bar{t}^{3d}_m$ to fit the real 3D template of the detected vehicle~$i$. This vector encodes the similarity between the detected vehicle and all the 3D templates  $\{\bar{t}^{3d}_m\}_{m \in \{1,..,M\}} $ of the 3D template dataset. 

At this stage of the approach, non-maximum suppression is performed to remove redundant detections. This provides a new set of $K'$ detections and associated attributes $\{B_{j},\bold{S}_j, \bold{V}_j,\bold{T}_j\}_{j \in \{1,..,K'\}}$.

\subsection{Deep MANTA Inference}
\label{Inference}

The inference step uses the Deep MANTA network outputs, the 3D shape dataset $\{\bar{\bold{S}}_m^{3d}\}_{m \in \{1,.., M\}}$ and the 3D template dataset $\{\bar{t}_m^{3d}\}_{m \in \{1,.., M\}}$ defined in~\ref{3dshapetemplate} to recover 3D information. Given a vehicle detection $j$ provided by the Deep MANTA network, the inference consists in two steps. In the first step, we choose the closest 3D template $c \in \{1,.., M\}$ in the 3D template dataset $\{\bar{t}_m^{3d}\}_{m \in \{1,.., M\}}$ using the template similarity $\bold{T}_j=\{r_m\}_{m \in \{1,.., M\}}$ returned by the network. For each sample $\bar{t}_m^{3d}$ of the 3D template dataset we apply the scaling transformation $r_m$. The resulting 3D templates are defined by $\{t_{m}^{3d}\}_{m \in \{1,.., M\}}$. The best 3D template $c$ is the one that minimizes the distance between $t_{m}^{3d}$ and $\bar{t}_m^{3d}$:
\[
c = \operatornamewithlimits{argmin}_{m \in \{1,.., M\}} d(\bar{t}_m^{3d},t_{m}^{3d}).
\]
In other words, the best 3D template is the one that is predicted closer to $(1,1,1)$ by the Deep MANTA network.

In the second step, 2D/3D matching is applied using 3D shape $\bar{\bold{S}}_{c}^{3d}$. It is rescaled to fit the 3D template $t_j=t_{c}^{3d}$. Then, a pose estimation algorithm is performed to match the rescaled 3D shape $\bar{\bold{S}}_c^{3d}$ with the 2D shape $\bold{S}_j$ using a standard 2D/3D matching~\cite{epnp}. This last step provides the 3D bounding box $B_j^{3d}$ and the 3D part coordinates $\bold{S}_j^{3d}$. The last block in Figure~\ref{fig:shapercnn} illustrates the inference step.

\section{Deep MANTA Training}

This section defines all the tasks of the MANTA network and the associated loss functions.
In the following, we consider three levels of refinement $l \in \{1,2,3\}$ and five functions to minimize: $\mathcal{L}_{rpn}$, $\mathcal{L}_{det}$, $\mathcal{L}_{parts}$, $\mathcal{L}_{vis}$ and $\mathcal{L}_{temp}$. $\mathcal{L}_{rpn}$ is the RPN loss function defined in~\cite{faster}. $\mathcal{L}_{det}$ is the detection loss function focusing on discriminating vehicle and background bounding box as well as regressing bounding boxes. $\mathcal{L}_{parts}$ is the loss corresponding to vehicle part localization. $\mathcal{L}_{vis}$ is the loss related to part visibility. $\mathcal{L}_{temp}$ is the loss related to template similarity. We use the Faster-RCNN framework~\cite{faster} based on RPN to learn the end-to-end MANTA model. Given an input image, the network joint optimization minimizes the global function:
\begin{equation}
\mathcal{L} = \mathcal{L}^{1} + \mathcal{L}^{2} + \mathcal{L}^{3} \notag
\end{equation}
with
\begin{align}
\mathcal{L}^1 &=  \mathcal{L}_{rpn}, \notag \\[8pt]
\mathcal{L}^2 &= \sum_i \mathcal{L}_{det}^2(i) + \mathcal{L}_{parts}^2(i), \notag \\
\mathcal{L}^3 &=  \sum_i  \mathcal{L}_{det}^3(i) + \mathcal{L}_{parts}^3(i) +  \mathcal{L}_{vis}(i) + \mathcal{L}_{temp}(i), \notag 
\end{align}
where $i$ is the index of a proposal object. These three losses correspond to the three levels of refinement of the Deep MANTA architecture: finer is the level, bigger is the amount of information learned.

\subsection{Many-task loss functions}
\label{manytask_training}
Here, we will detail the different task losses used in the global function presented above. In the following, each object proposal at each level of refinement $l$, is indexed by $i$ and it is represented by its box $B_{i,l}=(c_{x_{i,l}},c_{y_{i,l}},w_{i,l},h_{i,l})$. 
The closest ground-truth vehicle box $B$ to $B_{i,l}$ is selected. Associated ground-truth parts $\bold{S}$, ground-truth visibility $\bold{V}$ and ground-truth template $t$ are also selected (see section~\ref{Vehicle shape model}). We denote the standard log softmax loss as $P$ and the robust SmoothL1 loss defined in~\cite{fast} as $R$.

\textbf{Detection loss}. The object proposal $i$ at the refinement level $l$ is assigned to a class label $C_{i,l}$. $C_{i,l}$ is 1 if the object proposal is a vehicle and 0 otherwise. The classification criteria is the overlap between the box $B_{i,l}$ and the ground-truth box $B$. The predicted class returned by Deep MANTA network for the proposal is $C^*_{i,l}$. A target box regression vector  $\Delta_{i,l} = (\delta_x, \delta_y, \delta_w, \delta_h)$ is also defined as follows:
\begin{align}
\delta_x &= (c_{x_{i,l}}-c_x)/w &  \delta_w &= log(w_{i,l}/w)\notag \\ 
\delta_y &= (c_{y_{i,l}}-c_y)/h &  \delta_h &= log(h_{i,l}/h)  \notag
\end{align}
The predicted regression vector returned by Deep MANTA network is $\Delta_{i,l}^*$. The detection loss function is defined by:
\begin{align}
\mathcal{L}_{det}^l(i) & = \lambda_{cls} P(C_{i,l}^*,C_{i,l})+ \lambda_{reg} C_{i,l} R(\Delta_{i,l}^*- \Delta_{i,l}) \notag
\end{align}
with $\lambda_{cls}$ and $\lambda_{reg}$ the regularization parameters of box classification and box regression respectively. 
%To one level to another, box offsets are dynamically computed on proposals.

\textbf{Part loss.} Using the ground-truth parts $\bold{S}=(q_1,.., q_N)$ and the box $B_{i,l}$ associated to the object proposal $i$ at level~$l$, normalized vehicle parts $\bold{S}_{i,l} = (\bar q_1,.., \bar q_N)$ are computed as follows:
\[
\bar q_k = (\frac{u_k-c_{x_{i,l}}}{w_{i,l}} ,\frac{v_k-c_{y_{i,l}}}{h_{i,l}}).
\]
The predicted normalized parts are $S^*_{i,l}$. The part loss function is defined as:
\begin{align}
\mathcal{L}_{parts}^l(i) = \lambda_{parts} C_{i,l} R(\bold{S}_{i,l}^*- \bold{S}_{i,l}) \notag
\end{align}
with $\lambda_{parts}$ the regularization parameter of part loss.

\textbf{Visibility loss.} This loss is only optimized on the final level of refinement $l=3$. The ground-truth visibility vector $\bold{V}_{i} = \bold{V}$ is assigned to the object proposal~$i$. The predicted visibility vector is $\bold{V}^*_{i}$. The visibility loss function is defined as:
\begin{align}
\mathcal{L}_{vis}(i) = \lambda_{vis}  C_{i,3} P(\bold{V}_{i}^*,\bold{V}_{i}) \notag
\end{align}
with $\lambda_{vis}$ the regularization parameter of visibility loss.

\textbf{Template similarity loss.} This loss is only optimized on the final level of refinement $l=3$. Instead of directly optimizing the three dimensions of the 3D template $t$, we encode it as a vector $\bold{T}$ using the 3D template dataset as explained in~\ref{Deep MANTA network}. For training, the $log$ function is applied to each element of $\bold{T}$ for better normalization (similarity values are thus in $[-1,1]$). The ground-truth template similarity vector vector $\bold{T}_{i} = \bold{T}$ is assigned to the object proposal~$i$. The predicted template similarity vector is $\bold{T}^*_{i}$. The template similarity loss function is defined as:
\begin{align}
\mathcal{L}_{temp}(i) = \lambda_{temp} C_{i,3} R({\bold{T}_{i}^{*}} - \bold{T}_{i}) \notag
\end{align}
with $\lambda_{temp}$ the regularization parameter of template similarity loss. 

Notice that if the object proposal $i$ is not positive (\textit{i.e} $C_{i,l} = 0$) the loss functions associated to bounding box regression, part location, visibility and template similarity are null because it does not make sense to optimize vehicle properties on background regions.

\subsection{Semi-automatic annotation}
\label{semi}

\begin{figure*}[ht]
\center
\includegraphics[width=17cm]{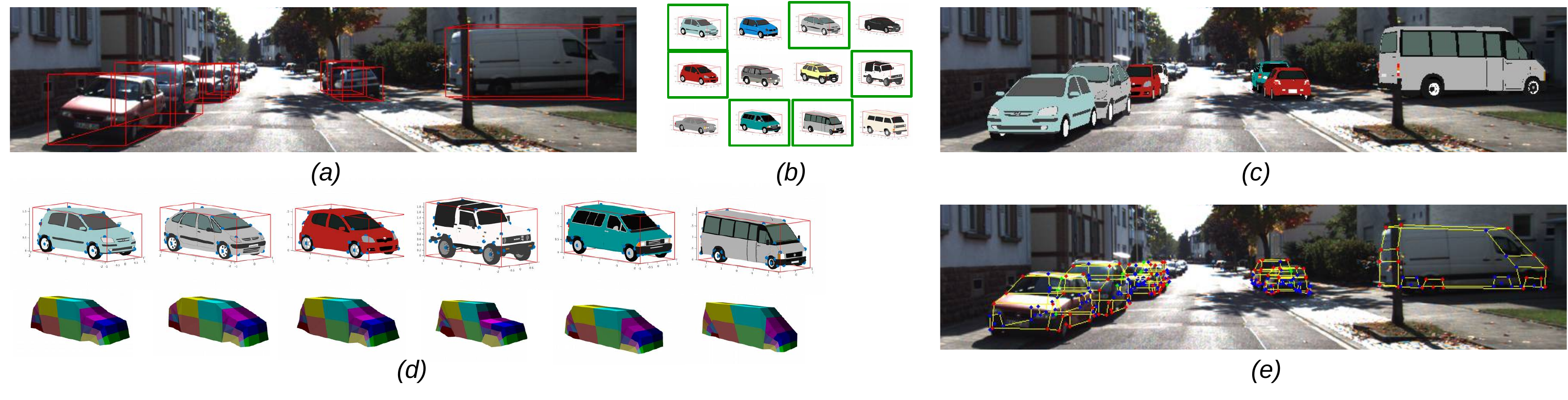}
\vspace{-0mm}
\caption{Semi-automatic annotation process. \textit{(a)} weak annotations on a real image (3D bounding box). \textit{(b)} best corresponding 3D models in green. \textit{(c)} projection of these 3D models in the image. \textit{(d)} corresponding mesh of visibility (each color represents a part). \textit{(e)} Final annotations (part localization and visibility). Red dots: visible parts, green dots: occluded parts, bleu dots: self-occluded parts. }
\label{fig:semi_img}
\end{figure*}

A semi-automatic annotation process is used to provide useful labels to train our Deep MANTA network (vehicles part coordinates, part visibility, 3D template). To perform the annotation process, we only need a weakly annotated real dataset providing 3D bounding boxes of vehicle and a 3D CAD dataset. For this purpose, we use a 3D CAD dataset composed of $M$ 3D car models. We manually annotate $N$ vertices on each 3D model. For each vehicle in the weakly annotated real dataset, we choose automatically the best corresponding 3D model in the 3D model dataset. This is done by choosing the 3D model which has its 3D bounding box closest to the real 3D vehicle bounding box in the image (in terms of 3D dimensions). 3D parts associated to the chosen CAD are projected onto the image to get 2D part coordinates. The visibility of each projected part is computed using a mesh of visibility. This mesh is a low resolution 3D model where each face is associated to an annotated vehicle 3D part. Figure~\ref{fig:semi_img} illustrates this process.

%%%%%%%%%%%%%%%%%%%%%%%%%%%%%%%%%%%%%%%%%%%%%%%%%%%
%%%%%%%%%%%%%%%%%%%%%%%%%%%%%%%%%%%%%%%%%%%%%%%%%%%\\
%%%%%%%%%%%%%%%%%%%%%%%%%%%%%%%%%%%%%%%%%%%%%%%%%%%\\
%%%%%%%%%%%%%%%%%%%%%%%%%%%%%%%%%%%%%%%%%%%%%%%%%%%\\
%%%%%%%%%%%%%%%%%%%%%%%%%%%%%%%%%%%%%%%%%%%%%%%%%%%\\
%%%%%%%%%%%%%%%%%%%%%%%%%%%%%%%%%%%%%%%%%%%%%%%%%%%\\
%%%%%%%%%%%%%%%%%%%%%%%%%%%%%%%%%%%%%%%%%%%%%%%%%%%\\

\section{Experiments}
\label{sec:experiments}

\begin{table*}[ht]
\centering
\resizebox{17cm}{!}{
\begin{tabular}{ccc|c|c|c|c|c|c|}

\cline{4-9} 
                     &                      & \multicolumn{1}{l|}{} & \multicolumn{3}{c|}{AP}                                               & \multicolumn{3}{c|}{AOS}  \\ \hline
\multicolumn{1}{|c|}{Method} & \multicolumn{1}{c|}{Type} & Time  & Easy & Moderate & Hard & Easy & Moderate  & Hard \\ \hline

\multicolumn{1}{|c|}{3DVP \cite{3dvp}}     			& \multicolumn{1}{c|}{Mono} & 40 s & 80.48 / \textcolor{white}{00}-\textcolor{white}{00} & 68.05 / \textcolor{white}{00}-\textcolor{white}{00} & 57.20 / \textcolor{white}{00}-\textcolor{white}{00} & 78.99 / \textcolor{white}{00}-\textcolor{white}{00} & 65.73 / \textcolor{white}{00}-\textcolor{white}{00} & 54.67 / \textcolor{white}{00}-\textcolor{white}{00} \\
\multicolumn{1}{|c|}{Faster-RCNN \cite{faster}}     	& \multicolumn{1}{c|}{Mono} & 2 s & 82.91 / \textcolor{white}{00}-\textcolor{white}{00} & 77.83 / \textcolor{white}{00}-\textcolor{white}{00} & 66.25 / \textcolor{white}{00}-\textcolor{white}{00} & \textcolor{white}{00}-\textcolor{white}{00} / \textcolor{white}{00}-\textcolor{white}{00}  & \textcolor{white}{00}-\textcolor{white}{00} / \textcolor{white}{00}-\textcolor{white}{00} & \textcolor{white}{00}-\textcolor{white}{00} / \textcolor{white}{00}-\textcolor{white}{00} \\ 
\multicolumn{1}{|c|}{SubCNN \cite{subcnn}}     		& \multicolumn{1}{c|}{Mono} & 2 s & 95.77 / \textcolor{white}{00}-\textcolor{white}{00} & 86.64 / \textcolor{white}{00}-\textcolor{white}{00} & 74.07 / \textcolor{white}{00}-\textcolor{white}{00} & 94.55 / \textcolor{white}{00}-\textcolor{white}{00} & 85.03 / \textcolor{white}{00}-\textcolor{white}{00} & 72.21 / \textcolor{white}{00}-\textcolor{white}{00}  \\
\multicolumn{1}{|c|}{3DOP \cite{3dop}}     			& \multicolumn{1}{c|}{Stereo} & 3 s & \textcolor{white}{00}-\textcolor{white}{00} / 94.49 & \textcolor{white}{00}-\textcolor{white}{00} / 89.65 & \textcolor{white}{00}-\textcolor{white}{00} / 80.97 & \textcolor{white}{00}-\textcolor{white}{00} / 92.98 & \textcolor{white}{00}-\textcolor{white}{00} / 87.34 & \textcolor{white}{00}-\textcolor{white}{00} / 78.24\\
\multicolumn{1}{|c|}{Mono3D \cite{mono3d}}     		& \multicolumn{1}{c|}{Mono} & 4.2 s & \textcolor{white}{00}-\textcolor{white}{00} / 95.75 & \textcolor{white}{00}-\textcolor{white}{00} / 90.01 & \textcolor{white}{00}-\textcolor{white}{00} / 80.66 & \textcolor{white}{00}-\textcolor{white}{00} / 93.70 & \textcolor{white}{00}-\textcolor{white}{00} / 87.61 & \textcolor{white}{00}-\textcolor{white}{00} / 78.00 \\ \hline
%\multicolumn{1}{|c|}{RPN DET sep}     		& \multicolumn{1}{c|}{Mono} & 0.7 s & \textcolor{white}{00}-\textcolor{white}{00} / 96.79 & \textcolor{white}{00}-\textcolor{white}{00} / 88.91 & \textcolor{white}{00}-\textcolor{white}{00} / 80.75 & \textcolor{white}{00}-\textcolor{white}{00} / 96.63 & \textcolor{white}{00}-\textcolor{white}{00} / 88.65 & \textcolor{white}{00}-\textcolor{white}{00} / 80.33 \\ \hline
%\multicolumn{1}{|c|}{Ours weights1}     		& \multicolumn{1}{c|}{Mono} & 0.7 s & \textcolor{white}{00}-\textcolor{white}{00} / 97.58 & \textcolor{white}{00}-\textcolor{white}{00} / 91.06 & \textcolor{white}{00}-\textcolor{white}{00} / 82.82 & \textcolor{white}{00}-\textcolor{white}{00} / 97.48 & \textcolor{white}{00}-\textcolor{white}{00} / 90.85 & \textcolor{white}{00}-\textcolor{white}{00} / 82.46 \\ \hline
%\multicolumn{1}{|c|}{Ours weights3}     		& \multicolumn{1}{c|}{Mono} & 0.7 s & \textcolor{white}{00}-\textcolor{white}{00} / 97.45 & \textcolor{white}{00}-\textcolor{white}{00} / 91.41 & \textcolor{white}{00}-\textcolor{white}{00} / 82.80 & \textcolor{white}{00}-\textcolor{white}{00} / 97.36 & \textcolor{white}{00}-\textcolor{white}{00} / 91.22 & \textcolor{white}{00}-\textcolor{white}{00} / 82.47 \\ \hline
\multicolumn{1}{|c|}{Ours GoogLenet}     				& \multicolumn{1}{c|}{Mono} & 0.7 s & \textbf{97.90} / \textbf{97.58} & 91.01 / 90.89 & \textbf{83.14} / 82.72& \textbf{97.60} / \textbf{97.44} & 90.66 / 90.66 & \textbf{82.66} / 82.35 \\ 
\multicolumn{1}{|c|}{Ours VGG16}     			& \multicolumn{1}{c|}{Mono} & 2 s & 97.45 / 97.2 & \textbf{91.47} / \textbf{91.85} & 81.79 / \textbf{85.15} & 97.10 / 97.09& \textbf{91.01} / \textbf{91.57} & 81.14 / \textbf{84.72}\\

\hline
\end{tabular}
}
\vspace*{2mm}
\caption{Results for 2D vehicle detection (AP) and orientation (AOS) on KITTI val sets. Results on the two validation sets: \textit{val1} / \textit{val2}.}
\label{AP_AOS_val}
\end{table*}

%%%%%%%%%%%%%%%%%%%%%%%%%%%%%%%%%%%%%%%%%%%
%%%%%%%%%%%%%%%%%%%%%%%%%%%%%%%%%%%%%%%%%%%

\begin{table*}[]
\centering

\begin{tabular}{c|c|c|c|l|c|c|}
\cline{2-7}
\multicolumn{1}{l|}{}&\multicolumn{3}{c|}{AP}             & \multicolumn{3}{c|}{AOS}                                                      \\ \cline{2-7} 
                         & Easy & Moderate & \multicolumn{1}{l|}{Hard} & Easy                     & \multicolumn{1}{l|}{Moderate} & \multicolumn{1}{l|}{Hard} \\ \hline
\multicolumn{1}{|c|}{LSVM-MDPM-sv \cite{Felzenszwalb10, Geiger11}} & 68.2 & 56.48 & 44.18 & \multicolumn{1}{c|}{67.27}  & 55.77 & 43.59                       \\
\multicolumn{1}{|c|}{ACF-SC \cite{cadena2015icra}} & 69.11 & 58.66 & 45.95                      & \multicolumn{1}{c|}{-} & - & -                       \\ 
\multicolumn{1}{|c|}{MDPM-un-BB \cite{Felzenszwalb10}} & 71.19 & 62.16 & 48.43                       & \multicolumn{1}{c|}{-} & - & -                   \\ 
\multicolumn{1}{|c|}{DPM-VOC+VP \cite{bojan15pami}} & 74.95 & 64.71 & 48.76                      & \multicolumn{1}{c|}{72.28} & 61.84 & 46.54                       \\ 
\multicolumn{1}{|c|}{OC-DPM \cite{bojan13cvpr}} &75.94 & 65.95 & 53.56                    & \multicolumn{1}{c|}{73.50} & 64.42 & 52.40                  \\ 
\multicolumn{1}{|c|}{SubCat \cite{subcat15}} & 84.14 & 75.46 & 59.71                      & \multicolumn{1}{c|}{83.41} & 74.42 & 58.83                     \\ 
\multicolumn{1}{|c|}{3DVP \cite{3dvp}} & 87.46 & 75.77 & 65.38                       & \multicolumn{1}{c|}{87.46} & 75.77 & 65.38                      \\ 
\multicolumn{1}{|c|}{AOG \cite{CarAOG_ECCV2014}} & 84.80 & 75.94 & 60.70                      & \multicolumn{1}{c|}{33.79} & 30.77 & 24.75                    \\ 
\multicolumn{1}{|c|}{Regionlets \cite{Regionlets-Relocalization}} & 84.75 & 76.45 & 59.70 & \multicolumn{1}{c|}{-} & - & - \\
\multicolumn{1}{|c|}{Faster R-CNN \cite{faster}} & 86.71 & 81.84 & 71.12                       & \multicolumn{1}{c|}{-} & - & -                     \\ 
\multicolumn{1}{|c|}{3DOP \cite{3dop} } & 93.04 & 88.64 & 79.10                      & \multicolumn{1}{c|}{91.44} & 86.10 & 76.52 \\
\multicolumn{1}{|c|}{Mono3D \cite{mono3d} } & 92.33 & 88.66 & 78.96                      & \multicolumn{1}{c|}{91.01} & 86.62 & 76.84\\  
\multicolumn{1}{|c|}{SDP + RPN \cite{sdp}} & 90.14  & 88.85  & 78.38                       & \multicolumn{1}{c|}{-} & -  & -                       \\
\multicolumn{1}{|c|}{MS-CNN \cite{mscnn} } & 90.03 & 89.02 & 76.11                      & \multicolumn{1}{c|}{-} & - & - \\ 
\multicolumn{1}{|c|}{SubCNN \cite{subcnn} } & 90.81 & 89.04 & 79.27                      & \multicolumn{1}{c|}{90.67} & 88.62 & 78.68 \\ \hline
%\multicolumn{1}{|c|}{Ours GoogLenet no vis} & 92.13  & 89.86  & 80.55                       & \multicolumn{1}{c|}{92.05} & 89.67  & 80.33                      \\
\multicolumn{1}{|c|}{Ours Googlenet} & 95.77 & 90.03  & 80.62                       & \multicolumn{1}{c|}{95.72} & 89.86  & 80.39                       \\
\multicolumn{1}{|c|}{Ours VGG16} & \textbf{96.40}  & \textbf{90.10}  & \textbf{80.79}                       & \multicolumn{1}{c|}{\textbf{96.32}} & \textbf{89.91}  & \textbf{80.55}                       \\

\hline

\end{tabular}
\vspace*{2mm}
\caption{Results for 2D vehicle detection (AP) and orientation (AOS) on the KITTI test set.}
\label{AP_AOS_test}
\end{table*}

\begin{table*}[]
\centering
\begin{tabular}{ccc|c|c|c|c|c|c|}
\cline{4-9}                                          &                          &     & \multicolumn{3}{c|}{AP} & \multicolumn{3}{c|}{AOS} \\ \hline
\multicolumn{1}{|c|}{Methode}             & \multicolumn{1}{c|}{Refinement} & ROI Pooling on & Easy  & Moderate & Hard & Easy  & Moderate  & Hard \\ \hline
\multicolumn{1}{|c|}{\multirow{3}{*}{Deep MANTA}} & \multicolumn{1}{c|}{No} & conv5  & 80.64 & 62.45 & 53.86 & 79.68 & 61.49 & 52.58    \\
\multicolumn{1}{|c|}{}                    & \multicolumn{1}{c|}{No} & conv1 & 95.19 & 86.85 & 78.62 & 94.98 & 86.52 & 78.05    \\ \cline{2-9} 
\multicolumn{1}{|c|}{}                    & \multicolumn{1}{c|}{Yes} & conv1  & \textbf{97.58} & \textbf{90.89} & \textbf{82.72} & \textbf{97.44} & \textbf{90.66} & \textbf{82.35}    \\ \hline
\end{tabular}
\vspace*{2mm}
\caption{Coarse-to-fine comparison for 2D vehicle detection (AP) and orientation estimation (AOS) on the validation set \textit{val2}. These experiments show the importance of the refinement step as well as the influence of the feature maps chosen for region extraction.}
\label{ctf}
\end{table*}
%%%%%%%%%%%%%%%%%%%%%%%%%%%%%%%%%%%%%%%%%%%%%%%%%%%
%%%%%%%%%%%%%%%%%%%%%%%%%%%%%%%%%%%%%%%%%%%%%%%%%%%\\
%%%%%%%%%%%%%%%%%%%%%%%%%%%%%%%%%%%%%%%%%%%%%%%%%%%\\
%%%%%%%%%%%%%%%%%%%%%%%%%%%%%%%%%%%%%%%%%%%%%%%%%%%\\
%%%%%%%%%%%%%%%%%%%%%%%%%%%%%%%%%%%%%%%%%%%%%%%%%%%\\
%%%%%%%%%%%%%%%%%%%%%%%%%%%%%%%%%%%%%%%%%%%%%%%%%%%\\
%%%%%%%%%%%%%%%%%%%%%%%%%%%%%%%%%%%%%%%%%%%%%%%%%%%\\

In this section, we evaluate the proposed approach on the challenging KITTI object detection benchmark dedicated to autonomous driving~\cite{kitti}. This dataset is composed of 7481 training images and 7518 testing images. The calibration matrix is given. Since ground truth annotations for the testing set are not released, we use train/validation splits from the training set to validate our method. To compare our approach to other state-of-the-art methods, we use two train/val splits: \textit{val1} used by~\cite{subcnn,3dvp} and \textit{val2} used by~\cite{3dop,mono3d}. This is a means to compare our approach to these methods for tasks which are not initially evaluated on the KITTI benchmark. We use the 3D CAD dataset provided by~\cite{cuboid,MTurkers} composed of $M=103$ 3D vehicle models for semi-automatic annotation. We annotate $N=36$ vehicle parts on each 3D model. We train the Deep MANTA using the GoogLenet~\cite{gn} and the VGG16~\cite{vgg} architectures with the standard stochastic gradient descent optimization. The Deep MANTA is initialized using pre-trained weights learned on ImageNet. We use 7 aspect ratios and 10 scales for the RPN providing 70 anchors at each feature map location as proposed by~\cite{subcnn}. During training, an object proposal is considered positive if its overlap with a ground-truth box is greater than 0.7. For experiments, all regularization parameters $\lambda$ are set to 1 except for the part localization task where $\lambda_{parts} = 3$. The choice of these parameters are discussed at the end of this section.

We present results for several tasks: 2D vehicle detection and orientation, 3D localization, 2D part localization, part visibility and 3D template prediction. In all presented results, we use 200 object proposals and an overlapping threshold of 0.5 for non-maximum suppression. Results are presented for three levels of difficulty (Easy, Moderate and Hard) as proposed by the KITTI Benchmark~\cite{kitti}.

\textbf{2D vehicle detection and orientation}. We use mean Average Precision (mAP) with overlapping criteria of 0.7 to evaluate 2D vehicle detection. We use average orientation similarity (AOS) to evaluate vehicle orientation as proposed by the KITTI Benchmark~\cite{kitti}. Table~\ref{AP_AOS_val} shows results for these two tasks on the two train/val splits. Table~\ref{AP_AOS_test} shows results on the KITTI testing set. We can see that our method outperforms others for the two tasks on the two train/val split as well as on the test set. In addition, our approach is less time consuming. This is due to the resolution of the input image. Many state-of-the-art object proposal based approaches~\cite{subcnn,mono3d, 3dop} upscale the input image by a factor of 3 on the KITTI dataset. This is done to not lose information on spatially reduced feature maps. Our coarse-to-fine approach overcomes this loss of information and that allows to give an input image at initial resolution.  The coarse-to-fine architecture of the Deep MANTA is also evaluated and results are shown in Table~\ref{ctf}. We compare the presented Deep MANTA to two other networks. The first line is a network which does not use refinement steps and where pooling regions are extracted on the feature map at the 5th level of convolution (as the original Faster-RCNN~\cite{faster}). The second line is a network without refinement steps and where pooling regions are extracted at the first level of convolution. We can see that extracting regions on the first convolution level clearly boosts detection and orientation score (around 24\% up for moderate). The last line is the presented Deep MANTA architecture (with refinement step and regions extracted on the first convolution maps). These results shows that the coarse-to-fine architecture increases detection and orientation estimation (around 4\% up for moderate).

\begin{figure}[ht]
\center
\includegraphics[width=8.4cm]{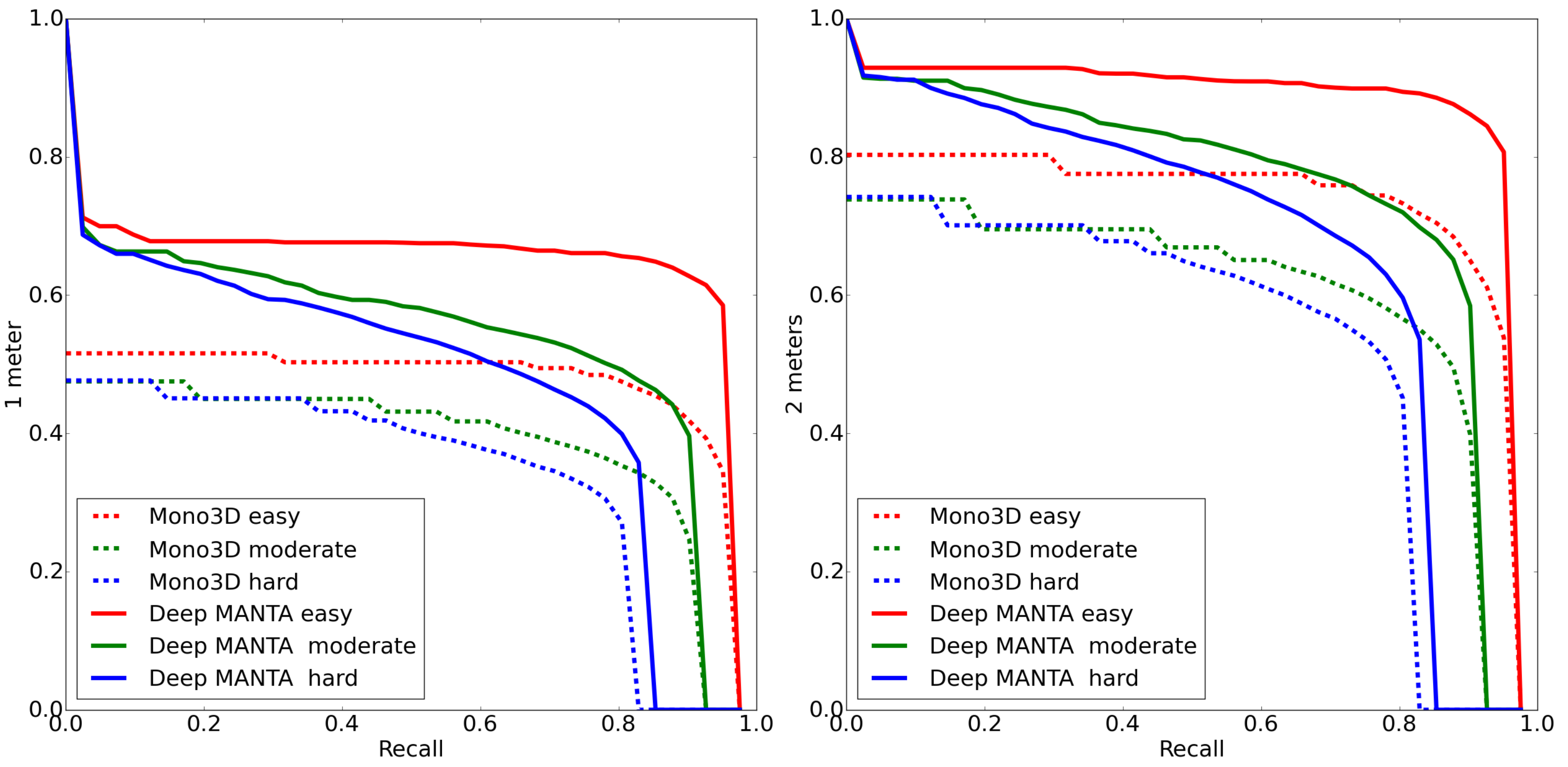}
\vspace{-3mm}
\caption{Recall/3D localization precision curves for 1 meter (left) and 2 meters (right) precision on the \textit{val2} used by Mono3D~\cite{mono3d}.}
\label{fig:curves}
\end{figure}

\textbf{3D localization}. We use Average Localization Precision (ALP) metric proposed by~\cite{3dvp}. It consists in replacing orientation similarity in AOS with localization precision. A 3D location is correct if its distance from the ground truth 3D location is smaller than a threshold. Table~\ref{ALP_val} presents results on the two train/val splits for a threshold distance of 1 meter and 2 meters. Our Deep MANTA approach clearly outperforms other monocular approaches~\cite{mono3d,3dvp} for the 3D localization task (around 16\% up compared to Mono3D~\cite{mono3d}). Figure~\ref{fig:curves} shows recall/3D localization precision curves of Deep MANTA and Mono3D~\cite{mono3d}. Compared to 3DOP~\cite{3dop}, which uses stereo information, the Deep MANTA performances are equivalent at a threshold error distance of 2 meters but less accurate at 1 meter: Deep MANTA only uses a single image contrarily to the 3DOP approach which uses disparity information.
\begin{table*}[]
\centering
\resizebox{17cm}{!}{
\begin{tabular}{ccc|c|c|c|c|c|c|}
\cline{4-9}                      &                      & \multicolumn{1}{l|}{} & \multicolumn{3}{c|}{1 meter}                                               & \multicolumn{3}{c|}{2 meters}  \\ \hline
\multicolumn{1}{|c|}{Method} & \multicolumn{1}{c|}{Type} & Time  & Easy & Moderate & Hard & Easy & Moderate  & Hard \\ \hline
\multicolumn{1}{|c|}{3DVP \cite{3dvp}}     			& \multicolumn{1}{c|}{Mono} & 40 s & 45.61 / \textcolor{white}{00}-\textcolor{white}{00} & 34.28 / \textcolor{white}{00}-\textcolor{white}{00} & 27.72 / \textcolor{white}{00}-\textcolor{white}{00} & 65.73 / \textcolor{white}{00}-\textcolor{white}{00} & 54.60 / \textcolor{white}{00}-\textcolor{white}{00} & 45.62 / \textcolor{white}{00}-\textcolor{white}{00}\\ 
\multicolumn{1}{|c|}{3DOP \cite{3dop}}     			& \multicolumn{1}{c|}{Stereo} & 3 s & \textcolor{white}{00}-\textcolor{white}{00} / \textbf{81.97} & \textcolor{white}{00}-\textcolor{white}{00} / \textbf{68.15} & \textcolor{white}{00}-\textcolor{white}{00} / \textbf{59.85} & \textcolor{white}{00}-\textcolor{white}{00} / \textbf{91.46} & \textcolor{white}{00}-\textcolor{white}{00} / \textbf{81.63} & \textcolor{white}{00}-\textcolor{white}{00} / \textbf{72.97}\\
\multicolumn{1}{|c|}{Mono3D \cite{mono3d}}     		& \multicolumn{1}{c|}{Mono} & 4.2 s & \textcolor{white}{00}-\textcolor{white}{00} / 48.31 & \textcolor{white}{00}-\textcolor{white}{00} / 38.98 & \textcolor{white}{00}-\textcolor{white}{00} / 34.25 & \textcolor{white}{00}-\textcolor{white}{00} / 74.77 & \textcolor{white}{00}-\textcolor{white}{00} / 60.91 & \textcolor{white}{00}-\textcolor{white}{00} / 54.24\\
\hline
%\multicolumn{1}{|c|}{RPN DET sep}     		& \multicolumn{1}{c|}{Mono} & 0.7 s & \textcolor{white}{00}-\textcolor{white}{00} / 71.24 & \textcolor{white}{00}-\textcolor{white}{00} / 54.75 & \textcolor{white}{00}-\textcolor{white}{00} / 47.30 & \textcolor{white}{00}-\textcolor{white}{00} / 90.14 & \textcolor{white}{00}-\textcolor{white}{00} / 74.30 & \textcolor{white}{00}-\textcolor{white}{00} / 65.33 \\ \hline
%\multicolumn{1}{|c|}{Ours weights1}     		& \multicolumn{1}{c|}{Mono} & 0.7 s & \textcolor{white}{00}-\textcolor{white}{00} / 62.10 & \textcolor{white}{00}-\textcolor{white}{00} / 49.11 & \textcolor{white}{00}-\textcolor{white}{00} / 42.52 & \textcolor{white}{00}-\textcolor{white}{00} / 88.80 & \textcolor{white}{00}-\textcolor{white}{00} / 73.38 & \textcolor{white}{00}-\textcolor{white}{00} / 64.32 \\ \hline
%\multicolumn{1}{|c|}{Ours weights3}     		& \multicolumn{1}{c|}{Mono} & 0.7 s & \textcolor{white}{00}-\textcolor{white}{00} / 66.90 & \textcolor{white}{00}-\textcolor{white}{00} / 53.69 & \textcolor{white}{00}-\textcolor{white}{00} / 46.49 & \textcolor{white}{00}-\textcolor{white}{00} / 90.45 & \textcolor{white}{00}-\textcolor{white}{00} / 76.28 & \textcolor{white}{00}-\textcolor{white}{00} / 67.01 \\ \hline
\multicolumn{1}{|c|}{Ours GoogLenet}     				& \multicolumn{1}{c|}{Mono} & 0.7 s &  \textbf{70.90} / 65.71& \textbf{58.05} / 53.79& \textbf{49.00} / 47.21 & \textbf{90.12} / 89.29 & \textbf{77.02} / 75.92 & \textbf{66.09} / 67.28\\ 
\multicolumn{1}{|c|}{Ours VGG16}     			& \multicolumn{1}{c|}{Mono} & 2 s & 66.88 / 69.72  & 53.17 / 54.44 & 44.40 / 47.77 & 88.32 / 91.01 & 74.31 / 76.38 & 63.62 / 67.77\\
\hline
\end{tabular}
}
\vspace*{2mm}
\caption{3D localization accuracy (ALP) on KITTI val sets for 1 meter and 2 meters precision. Results on the two validation sets: \textit{val1} / \textit{val2}.}
\label{ALP_val}
\end{table*}

\textbf{3D template, part localization and visibility}. 
We also evaluate the precision of part localization, part visibility classification accuracy as well as 3D template prediction. Given a correct detection, we use the following three metrics. For part localization, a part is considered well localized if the normalized distance to the ground-truth part is less than a threshold (20 pixels). Distances are normalized using a fixed bounding box height (155 pixels) as proposed by~\cite{Zia3}. The visibility metric is the accuracy over the four visibility classes. Finally, we evaluate 3D template prediction by comparing the three predicted dimensions $(w,h,l)$ to the ground-truth 3D box dimensions $(w_{gt},h_{gt},l_{gt})$ provided by KITTI. A 3D template $(w,h,l)$ is considered correct if $\lvert {{{w_{gt} - w} \over w_{gt}}} \rvert < 0.2$ and $\lvert {{{h_{gt} - h} \over h_{gt}}} \rvert < 0.2$ and $\lvert {{{l_{gt} - l} \over l_{gt}}} \rvert < 0.2$. Table \ref{p_v_t} shows the good performances for these tasks.

\begin{table}[ht]
\centering
\begin{tabular}{|c|c|c|c|}
\hline
Metric & Easy & Moderate & Hard   \\ \hline
Part localization   & 97.54  & 90.79  & 82.64 \\
Part visibility     & 92.48  & 85.08  & 76.90  \\ 
3D template     & 94.04  & 86.62  & 78.72 \\ \hline
\end{tabular}
\vspace*{3mm}
\caption{Part localization, part visibility, 3D template evaluation on the validation set \textit{val2}.}
\label{p_v_t}
\end{table}

\textbf{Many-task and regularization parameters}. Table~\ref{medley} shows results with different sets of regularization parameters. These results also aim to compare performances of the Deep MANTA approach with networks optimized on fewer tasks. In Table~\ref{medley}, D corresponds to the detection task, P to the part localization task, V to the part visibility task and T to the template similarity task. With these notations, the first line of Table~\ref{medley} is the Deep MANTA trained only on the detection task ($\lambda_{parts} = \lambda_{vis} = \lambda_{temp} = 0$). As part localization and template similarity are not trained, orientation and 3D localization cannot be predicted in this case. The second line is the Deep MANTA trained without the visibility task ($\lambda_{vis} = 0$) and with $\lambda_{parts} = 3$. The third line is the complete Deep MANTA (all tasks) but with the regularization parameter associated to part localization $\lambda_{parts} = 1$. Finally, the last line is the Deep MANTA with $\lambda_{parts} = 3$ (the one presented in all above results). These results are interesting for several reasons. First, we can see that increasing the number of learned tasks (\textit{i.e} enriching the vehicle description) does not significantly affect performances (it is slightly higher for detection and orientation accuracy but slightly lower on 3D localization). That proves the relevance of the Many-Task concept: a neural network is able to learn one feature representation which can be used to predict many tasks. Secondly, we can see that the parameter $\lambda_{parts}$ is very important for 3D localization. Learning the Deep MANTA with $\lambda_{parts} = 3$ improves the 3D localization by  6\% for 1 meter distance precision.       
\begin{table}[ht]
\centering
\begin{tabular}{cccccc}
\multicolumn{1}{l}{}     & \multicolumn{1}{l}{}    & \multicolumn{1}{l}{}     & \multicolumn{1}{l}{}         & \multicolumn{1}{l}{}          \\ \cline{2-5} 
\multicolumn{1}{c|}{}    & \multicolumn{1}{c|}{AP} & \multicolumn{1}{c||}{AOS} &  \multicolumn{1}{c|}{1 m} & \multicolumn{1}{c|}{2 m} \\ \hline
\multicolumn{1}{|c|}{D} & \multicolumn{1}{c|}{89.86}  & \multicolumn{1}{c||}{-}   & \multicolumn{1}{c|}{-}       & \multicolumn{1}{c|}{-}        \\
\multicolumn{1}{|c|}{DPT / $\lambda_{parts} = 3$} & \multicolumn{1}{c|}{89.73}  & \multicolumn{1}{c||}{89.39}  & \multicolumn{1}{c|}{\textbf{58.37}}       & \multicolumn{1}{c|}{\textbf{78.11}}        \\ 
\multicolumn{1}{|c|}{DPVT / $\lambda_{parts} = 1$} & \multicolumn{1}{c|}{89.58}  & \multicolumn{1}{c||}{89.27}  & \multicolumn{1}{c|}{51.47}       & \multicolumn{1}{c|}{73.93}        \\ 
\multicolumn{1}{|c|}{DPVT / $\lambda_{parts} = 3$} & \multicolumn{1}{c|}{\textbf{90.54}}  & \multicolumn{1}{c||}{\textbf{90.23}}  & \multicolumn{1}{c|}{57.44} & \multicolumn{1}{c|}{77.58} \\ \hline
\end{tabular}
\vspace{3mm}
\caption{The influence of the amount of tasks learned as well as different regularization parameters. This table gives results for vehicle detection (AP), orientation (AOS), and 3D localization for 1 meter and 2 meters precision (ALP). Given results are averaged over the two validation sets and over the three levels of difficulty (Easy, Moderate, Hard). See text for details.}
\label{medley}
\end{table}
\section{Conclusion}
\label{sec:conclusion}
To conclude, we propose a new approach for joint 2D and 3D vehicle analysis from monocular image. It is based on the Many-task CNN (Deep MANTA) which proposes accurate 2D vehicle bounding boxes using multiple refinement steps. The MANTA architecture also provides vehicle part coordinates (even if these parts are hidden), part visibility and 3D template for each detection. These fine features are then used to recover vehicle orientation and 3D localization using robust 2D/3D point matching. Our approach outperforms state-of-the-art methods for vehicle detection and fine orientation estimation and clearly increases vehicle 3D localization compared to monocular approaches. One perspective is to adapt this framework to other rigid objects and build a multi-class Deep MANTA network.

{\small
\bibliographystyle{ieee}
\bibliography{egbib}
}

\end{document}